\documentclass[10pt,twocolumn,letterpaper]{article}

\usepackage[Final]{iccv}      

\definecolor{iccvblue}{rgb}{0.21,0.49,0.74}
\usepackage[pagebackref,breaklinks,colorlinks,allcolors=iccvblue]{hyperref}

\usepackage{multirow}
\usepackage{booktabs} 
\usepackage{graphicx}
\usepackage[accsupp]{axessibility}

\def\paperID{*****} 
\def\confName{ICCV}
\def\confYear{2025}

\title{CTFlow: Video-Inspired Latent Flow Matching for 3D CT Synthesis}

\author{Jiayi Wang\\
Friedrich–Alexander University Erlangen\\
Nürnberg, DE\\
{\tt\small jiayi.w.wang@fau.de}
\and
Hadrien Reynaud\\
Friedrich–Alexander University Erlangen\\
Nürnberg, DE\\
{\tt\small hadrien.reynaud@fau.de}
\and
Franciskus Xaverius Erick\\
Friedrich–Alexander University Erlangen\\
Nürnberg, DE\\
{\tt\small franciskus.erick@fau.de}
\and
Bernhard Kainz\\
Department of Computing, Imperial College London\\
London, UK\\
Friedrich–Alexander University Erlangen\\
Nürnberg, DE\\
{\tt\small b.kainz@imperial.ac.uk}
}

\usepackage[acronym]{glossaries}
\newacronym{cdm}{CDM}{Cascaded Diffusion Model}
\newacronym{ct}{CT}{Computed Tomography}
\newacronym{ddim}{DDIM}{Denoising Diffusion Implicit Model}
\newacronym[longplural={Denoising Diffusion Probabilistic Models},plural={DDPMs}]{ddpm}{DDPM}{Denoising Diffusion Probabilistic Model}
\newacronym{dnn}{DNN}{Deep Neural Network}
\newacronym{ed}{ED}{End-Diastolic}
\newacronym{es}{ES}{End-Systolic}
\newacronym{fid}{FID}{Fréchet Inception Distance}
\newacronym{fvd}{FVD}{Fréchet Video Distance}
\newacronym{fps}{fps}{frames per second}
\newacronym[longplural={Generative Adversarial Networks},plural={GANs}]{gan}{GAN}{Generative Adversarial Network}
\newacronym{is}{IS}{Inception Score}
\newacronym{ldm}{LDM}{Latent Diffusion Model}
\newacronym{lidm}{LIDM}{Latent Image Diffusion Model}
\newacronym{lvdm}{LVDM}{Latent Video Diffusion Model}
\newacronym{lpips}{LPIPS}{Perceptual Similarity}
\newacronym{mae}{MAE}{Mean Absolute Error}
\newacronym{mse}{MSE}{Mean Squared Error}
\newacronym{ssim}{SSIM}{Structural Similarity Index}
\newacronym{psnr}{PSNR}{Peak Signal-to-Noise Ratio}
\newacronym{flops}{FLOPs}{Floating-point Operations per Second}
\newacronym{r2}{$R^2$}{Coefficient of Determination}
\newacronym{rmse}{RMSE}{Root mean square error}
\newacronym{vae}{VAE}{Variational Auto-Encoder}
\newacronym{lifm}{LIFM}{Latent Image Flow Matching}
\newacronym{lvfm}{LVFM}{Latent Video Flow Matching}
\newacronym{avae}{A-VAE}{Adversarial Variational Auto-Encoder}
\newacronym{cfg}{CFG}{Classifier-Free Guidance}
\newacronym{reid}{ReId}{Re-Identification}
\newacronym{llm}{LLM}{Large Language Model}

\begin{document}
\maketitle

\begin{abstract}
Generative modelling of entire CT volumes conditioned on clinical reports has the potential to accelerate research through data augmentation, privacy-preserving synthesis and reducing regulator-constraints on patient data while preserving diagnostic signals.
With the recent release of CT-RATE, a large-scale collection of 3D CT volumes paired with their respective clinical reports, training large text-conditioned CT volume generation models has become achievable.
In this work, we introduce CTFlow, a 0.5B latent flow matching transformer model, conditioned on clinical reports. We leverage the A-VAE from FLUX to define our latent space, and rely on the CT-Clip text encoder to encode the clinical reports.
To generate consistent whole CT volumes while keeping the memory constraints tractable, we rely on a custom autoregressive approach, where the model predicts the first sequence of slices of the volume from text-only, and then relies on the previously generated sequence of slices and the text, to predict the following sequence.
We evaluate our results against state-of-the-art generative CT model, and demonstrate the superiority of our approach in terms of temporal coherence, image diversity and text-image alignment, with FID, FVD, IS scores and CLIP score.
\end{abstract}    
\section{Introduction}
\label{sec:intro}

Medical imaging analysis has witnessed substantial progress with the advancement of artificial intelligence and deep learning. These technologies are set to significantly improve various clinical tasks, including diagnostic accuracy and treatment planning workflows \cite{abdialkareemalyasseri2022review,litjens2016deep,mansouri2021deep}. However, critical challenges remain. Most current data-driven deep learning models depend heavily on real-world medical datasets, raising serious privacy concerns due to the sensitive nature of patient data and the strict ethical and regulatory constraints under which such data must be handled \cite{kaissis2021end-to-end,paul2023digitization,salimans2022progressive}.

To address these limitations, synthetic data has emerged as a promising solution for enhancing rare disease signals and augmenting limited datasets without relying entirely on real patient records. In particular, with the rapid development of \glspl{llm}, the generation of text-conditional medical images has become one of the most promising approaches to synthesize medical data \cite{xu2024medsyn,ethemhamamci2025generatect,bluethgen2024vision--language}. However, several critical issues in this domain remain unresolved.

One of the primary challenges lies in generating high-resolution 3D \gls{ct} volumes, which can span more than 600 slices along the axial dimension, with a resolution of $512\times512$ for each slice. Existing 3D generative frameworks struggle with such high-resolution outputs due to the immense memory requirements. To alleviate this, several prior works \cite{chen20242,ethemhamamci2025generatect,zhang2022bridging} have proposed hybrid architectures that combine 2D and 3D components. These models can be interpreted as multi-frame generation pipelines. However, similar to challenges in video generation, such two-stage super-resolution approaches often suffer from spatial discontinuities and grid-like artifacts~\cite{ethemhamamci2025generatect}.

Inspired by recent advances in autoregressive video generation~\cite{deng2024autoregressive}, we introduce CTFlow, a novel text-conditioned framework for high-resolution 3D \gls{ct} volume generation in latent space, using a long-range autoregressive strategy. 
Our framework consists of three components: an \gls{avae}, a CLIP-encoder, and a latent flow matching model.
We leverage the open-source high-performing FLUX~\cite{blackforest-flux1} \gls{avae} to define our latent space, as it was shown to perform well on medical imaging tasks \cite{reynaud2025echoflow}. FLUX, like other \glspl{avae}, relies on a combination of losses to ensure the best possible reconstruction quality, including an adversarial loss. This combination of losses preserves both anatomical fidelity and semantic realism.
For text encoding, we rely on CT-CLIP~\cite{ethemhamamci2024developing}, a CLIP~\cite{radford2021learning} model that was trained on the CT-RATE dataset, and thus acts like an in-domain expert compared to a general CLIP model.
On top of the FLUX-defined latent space, and conditioned on the CT-CLIP embeddings, we develop a \gls{lvfm} model, inspired by~\cite{reynaud2025echoflow}. Our model is trained as a conditional latent flow matching model, generating one sequence of latent slices at a time, conditioned on the directly preceding sequence of latent slices.
This conditioning allows us to sample the model auto-regressively during inference, by giving the previous latent sequence of slices to the model, and getting the model to generate the next sequence. By repeating this operation many times, while conditioning on the same radiology report, we can generate a consistent whole \gls{ct} sequence and thus a \gls{ct} volume.
To generate the starting slices, while training the model, we condition it on a sequence of latent black slices, akin to a ``start of sequence'' token in \glspl{llm}. We also train the model to output latent white slices when it reaches the end of the \gls{ct}-volume, and identify these slices as ``end of sequence'' slices, to stop the autoregressive generation.
This strategy is inspired by autoregressive pipelines in video generation such as SkyReel-V2~\cite{chen2025skyreels-v2}, which demonstrate that smaller temporal video sequences improve batch efficiency and generation stability.
We evaluate our model on a next-sequence generation task and on a whole-\gls{ct} generation task in the VLM3D challenge (\url{https://vlm3dchallenge.com/}), using generative metrics like \gls{fvd}, \gls{fid}, CLIP score and \gls{is}. 

\vspace{\baselineskip}

\noindent Our \textbf{contributions} can be summarized as follows.
\begin{enumerate}
    \item We introduce a novel perspective for 3D \gls{ct}  generation, by treating the volumetric data as sequences of slices, similar to how we would process a video. This enables  autoregressive generation techniques for high-resolution 3D medical imaging, addressing the challenges of memory efficiency and spatial (\emph{i.e.} axial) coherence.
    \item We provide a state-of-the-art latent flow matching model trained for 5,000+ GPU hours, allowing high quality next-sequence prediction and autoregressive whole-\gls{ct} generation. This design significantly improves training stability and scalability for whole-volume synthesis.
    \item We demonstrate that conditioning the generation of each latent sub-video clip on previously generated video clips effectively solves structural consistency across the volume, mitigating discontinuities and artifacts, common in established two-stage super-resolution approaches. 
\end{enumerate}

\section{Related Works}
\label{sec:related}

\textbf{Synthetic Generative Models.}
Video synthesis in computer vision has gained significant attention in recent years. \glspl{gan}~\cite{goodfellow2014generative} and \glspl{vae}~\cite{pkingma2014auto-encoding} have been widely adopted in this area. Diffusion models~\cite{ho2022video} have shown promising performance, capable of generating not only low temporal and spatial resolution sequences but also high-definition videos conditioned on text \cite{ho2022imagen, khachatryan2023text2video-zero,singer2022make-a-video}. To alleviate the high computational cost of these models, \glspl{ldm}~\cite{rombach2022high-resolution} were introduced for image and video generation \cite{he2023latent,blattmann2023align,reynaud2024echonet-synthetic}, converting pre-trained image \glspl{ldm} into video generators by incorporating temporal layers. Furthermore, the \gls{lvdm} \cite{he2023latent} leverages a hierarchical architecture to enable long video generation through a secondary model. Recently, flow matching has shown promises of a more efficient denoising formulation \cite{esser2024scaling,lee2024improving,lipman2023flow,rombach2022high-resolution} and some studies have begun to explore its application to improve video generation efficiency \cite{jin2024pyramidal,reynaud2025echoflow}.

\noindent\textbf{CT Image Generation.}
In this field, several works \cite{bluethgen2024vision--language,ethemhamamci2025generatect,xu2024medsyn,cho2024medisyn} have focused on text-conditioned \gls{ct} image generation. GenerateCT~\cite{ethemhamamci2025generatect} stands out as a particularly relevant method, employing a transformer-based model that extends 2D super-resolution networks to the 3D domain. However, the lack of continuity in the generated 3D volumes limits its practical applicability in clinical settings. MedSyn~\cite{xu2024medsyn} adopts a hierarchical UNet architecture to generate low-resolution 3D lung \gls{ct} volumes, followed by a super-resolution module to enhance output quality. MAISI~\cite{guo2025maisi} proposes a 3D latent diffusion model, pretrained on a large collection of unlabeled \gls{ct} volumes, and further conditioned by a ControlNet~\cite{zhang2023adding}, through segmentation masks of 128 classes.
Their generation pipeline relies on a 3D CT volumes \gls{avae}, allowing the model to generate high resolution whole-\glspl{ct}.

In this work, we depart from the traditional two-stage 3D volume generation framework by treating 3D volumes as sequences of 2D slices, \emph{i.e.}, videos. We introduce advanced flow-matching-based video generation techniques to improve generation efficiency, reduce memory requirements, and allow for a variable and unconstrained number of axial slices.

\section{Methodology}
\label{sec:method}

We train our model using clinical data and evaluate the results by comparing samples generated from synthetic data with real clinical validation data to assess their realism.

\begin{figure*}[t]
  \centering
  \includegraphics[width=\textwidth]{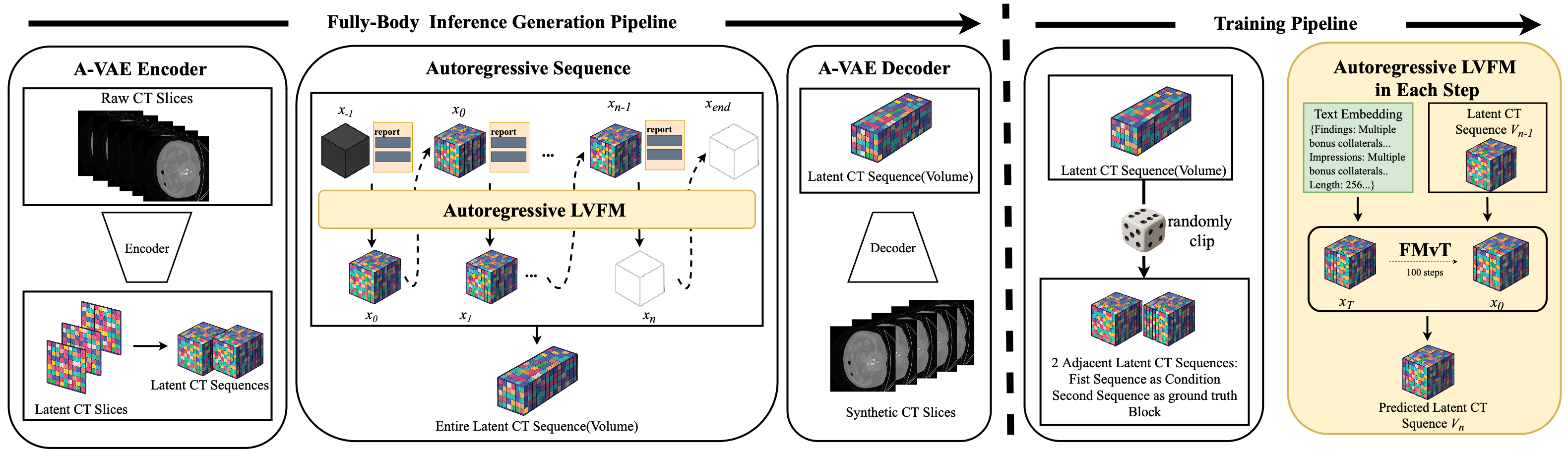}
  \caption{Our latent \gls{ct} fully-body inference pipeline and training pipeline. During inference, from left to right: The \gls{avae} encoder converts scaled \gls{ct} slices into latent \gls{ct} sequences; an autoregressive sequence is constructed starting from a ``start'' sequence; the latent volume is generated step-by-step via our \gls{lvfm} model. The \gls{avae} decoder reconstructs synthetic \gls{ct} slices. For each stage, inputs are shown at the top, processing modules in the middle, and outputs at the bottom. During training, we randomly sample two consecutive 16-slice sequences from the entire \gls{ct} volume to train the \gls{lvfm}. 
  }
  \label{fig:framework}
\end{figure*}

\subsection{Framework}
We rely on three core components: (1) an \gls{avae} to compress our \gls{ct} volumes into a latent space, (2) a text embedding model to project the clinical reports into fixed-size embeddings and (3) a latent sequence flow matching model to generate sequences of \gls{ct} slices.

\subsection{Adversarial Variational Auto-Encoder}

Training directly on full-resolution \gls{ct} volumes is computationally prohibitive, so we first project each slice onto a compact two-dimensional latent space with an \gls{avae}~\cite{rombach2022high-resolution}. Because the latent space is 2-dimensional, each latent pixel corresponds to a specific image patch, retaining the spatial and anatomical arrangement, while substantially reducing the memory and compute costs.

\glspl{vae} usually suffer from blurred reconstructions. \glspl{avae} solve this problem by using an adversarial loss. 
The resulting \gls{gan}-like content filling properties  ensure that our latent \gls{ct} volumes are decoded into high quality \gls{ct} volumes. Specifically, the discriminator model, used during the \gls{avae} training, pushes the decoder to restore the high-frequency details that the encoder might discard.



Nonetheless, \glspl{avae} come with the typical training instabilities of \glspl{gan}, and thus training a domain-specific \gls{avae} is not trivial. Fortunately, various well-performing \glspl{avae} are publicly available~\cite{rombach2022high-resolution,blackforest-flux1,stable-diffusion-3} and  work well on most type of images, including our \gls{ct} slices.

Starting from an input of shape $512 \times 512 \times 3$, the \glspl{avae} encoder produces a latent image of size $64 \times 64 \times 16$, corresponding to a $12\times$ reduction in the number of elements. Furthermore, as the input \gls{ct} slices are typically stored as \texttt{uint8} (1 byte per element), whereas the latent representations are stored as \texttt{float16} (2 bytes) or \texttt{float32} (4 bytes), the overall memory compression ratio becomes:
\[
\frac{512\times512\times3 \times 1}{64\times64\times16 \times 2} = 6\times \quad (\text{for float16}),
\]
and
\[
\frac{512\times512\times3 \times 1}{64\times64\times16 \times 4} = 3\times \quad (\text{for float32}).
\]
That is, after accounting for both spatial and channel changes and the increased data type size, the total storage required for the latents is reduced by $3\times$ to $6\times$ compared to the original data. This quantifies the trade-off between the substantial spatial compression and the partial offset from channel expansion and higher-precision data types.

It should be noted that increasing the number of latent channels $C$ linearly increases both the memory and compute requirements, while increasing spatial resolution ($H \times W$) results in a quadratic increase in computational complexity for transformer-based modules, due to the $O((H \times W)^2)$ self-attention mechanism. 


\subsection{CT-CLIP Embeddings}

We leverage the CT-CLIP text encoder~\cite{ethemhamamci2024developing,hamamci2024ct2rep}, which is based on the BiomedVLP-CXR-BERT-specialized architecture. This encoder is specifically designed to process radiology reports, which typically include sections such as ``Findings'' and ``Impressions'' as well as meta-information, such as the number of slices in the \gls{ct} volume (\emph{e.g.}, ``length of volume: 266''). The CT-CLIP encoder transforms these free-text radiology reports into dense semantic embeddings that capture clinically relevant information. These embeddings serve as conditioning signals for the generative model, enabling it to synthesize \gls{ct} volumes that are consistent with both the described findings and the expected scan properties. Notably, this enables the CLIP model to interpret the intended \gls{ct} volume length from natural language descriptions, and thus allowing our generator to produce the requested number of slices.

\subsection{Latent Flow Matching}

We adopt a flow matching framework to train our latent video generation model, which deterministically learns a continuous mapping from noise to the latent \gls{ct} manifold. Unlike diffusion-based models that rely on stochastic denoising, flow matching directly models a velocity field, guiding an interpolation trajectory between the prior distribution and real data.

Let $p_{\text{data}}$ denote the distribution of real latent \gls{ct} sequences and $p_{\text{prior}}$ a standard Gaussian prior. For training, we sample a pair $(x_0, x_1) \sim (p_{\text{data}}, p_{\text{prior}})$ and define an interpolated latent:

\begin{equation}
x_t = (1 - t)x_0 + t x_1, \quad t \in [0, 1].
\end{equation}

The corresponding ground-truth velocity is simply:

\begin{equation}
u(x_0, x_1) = \frac{d x_t}{d t} = x_1 - x_0.
\end{equation}

Our model learns a time-dependent velocity field $v_\theta(x_t, t)$ by minimizing the regression loss:

\begin{equation}
\mathcal{L}_{\text{FM}} = \mathbb{E}_{t, x_0, x_1} \left[ \left\| v_\theta(x_t, t) - (x_1 - x_0) \right\|^2 \right].
\end{equation}

During inference, we sample a noise latent $x_1 \sim p_{\text{prior}}$ and integrate the learned vector field backwards from $t=1$ to $t=0$ using the Euler method:

\begin{equation}
x_{t - \Delta t} = x_t - \Delta t \cdot v_\theta(x_t, t).
\end{equation}

In this work, we adopt a spatio-temporal transformer~\cite{vaswani2017attention} as the flow matching backbone $v_\theta$, following the architecture proposed in \cite{peebles2023scalable}. Each input latent sequence is divided into spatio-temporal patches, enriched with positional embeddings, and processed through a stack of alternating temporal and spatial attention layers. This design allows the model to capture fine-grained anatomical dynamics across slices.

We condition the model on prior latent sequences and text embeddings derived from the radiology reports, allowing it to generate consistent and clinically meaningful sequences.

\subsection{Training} 
During training, we pre-encode the whole-\gls{ct} volumes into sequences of latent slices using the \gls{avae} and pre-compute all the embeddings for the clinical reports.
We train the flow matching models by sampling two consecutive sequences of slices and the text embedding corresponding to that \gls{ct} volume. The model is trained to predict the second sequence while being conditioned on the first sequence and the text embedding.
To maximize the generalizability of the model, the sequences are sampled randomly, with no predefined spacing, \emph{i.e.} the starting slice index is not forced to be a multiple of the sequence length.
We train the model under two data sampling regimes. 
The first one is the naive sampling, where the sequences are sampled uniformly, giving each latent \gls{ct} slice the same probability of being sampled, including our start and end representations.
The second approach acknowledges that the beginning of the sequence, conditioned on the latent black sequence, is much more difficult to learn due to the less informative conditioning (\emph{i.e.}, text only). Therefore, we also train some models with a higher probability of sampling the first sequence.

\subsection{Inference and Autoregressive Sampling}

We evaluate our model under three distinct inference strategies, each designed to probe different aspects of its generative performance.

In the \textbf{fully-body} inference setting, the model auto-regressively generates the entire \gls{ct} volume conditioned only on the text embedding $c$. Starting from an initial all-black sequence $s_0$, the model iteratively predicts each subsequent latent sequence $s_{n+1}$ using the previous output $s_n$ and the text embedding:
\begin{equation}
s_{n+1} = G(s_n, c).
\end{equation}
This process continues until a white slice (serving as an end token) is generated. All predicted sequences are concatenated and decoded by the \gls{avae} to reconstruct the final \gls{ct} volume. This approach evaluates the model’s capacity for fully autonomous synthesis, given only clinical text, without any observed image context. It also directly tests its ability to learn long-range anatomical consistency across the whole scan.

In the \textbf{Ground-Truth-head} inference setting, the process is similar, except the first sequence $s_0$ is taken directly from ground-truth \gls{ct} scans rather than being synthesized:
\begin{equation}
s_{n+1} = G(s_n, c), \qquad s_0 = \text{ground-truth}.
\end{equation}
This provides plausible initial anatomical context from text alone, allowing us to focus on the model’s ability to generate realistic follow-up sequences when provided with an accurate anatomical top region. 
A secondary effect is that the model becomes less prone to compounding errors.

The \textbf{next-block} inference strategy further isolates local conditional generation ability. At each step, the model predicts $s_{n+1}$ using the \emph{ground-truth} previous sequence as context:
\begin{equation}
s_{n+1} = G(s^{\mathrm{GT}}_n, c), 
\end{equation}
where $s^{\mathrm{GT}}_n$ denotes the ground-truth sequence at location $n$. This bypasses the accumulation of errors during autoregressive generation, providing a direct measure of the model’s capacity to generate the next sequence, conditioned on accurate anatomical context and text. Furthermore, it offers a clean evaluation of the model’s conditional generative performance, disentangled from any bias introduced by previous prediction errors.

\section{Experiments}
\label{sec:experiments}

\subsection{Dataset and Implementation Details}

We conduct our experiments on the publicly available CT-RATE dataset~\cite{ethemhamamci2024developing}, which provides high-resolution 3D chest \gls{ct} volumes along with corresponding radiology reports. The dataset includes 47,100 training samples and 3,000 validation samples, covering a wide range of annotated \gls{ct} volumes with diverse pathological conditions. In our experiments, we use the latest fixed version of the training and validation subsets.

All \gls{ct} volumes slices are resampled to a fixed shape of $256 \times 256 \times 3$, encoded with the FLUX~\cite{blackforest-flux1} \gls{avae}. This results in latent volumes with shape $32 \times 32 \times 16 \times N$, where $N$ is the total number of slices in the original \gls{ct} volume.
Our networks are implemented using PyTorch, with a Spatio-Temporal Transformer backbone inspired from \cite{ma2024latte}. 
We fix the number of slices in a sequence to 16 and the patch size of the transformer model to $2 \times 2 \times 2$. We train three sizes of the transformer backbone: \textit{\underline{S}mall}, \textit{\underline{B}ase}, and \textit{\underline{L}arge}.


Training is performed using the latent \gls{ct} sequences, a learning rate of $2 \times 10^{-4}$, and a linear warm-up over the first 2000 steps. The model is trained on a distributed setup with 16 nodes, each with 4 NVIDIA H100 GPUs, using a batch size of 16 per GPU, and gradient accumulation of 2, for an effective batch size of 2048. We train for 100,000 training steps, adding up to 5,133 GPU-hours.

\begin{table}[ht]
  \centering
  \small
  \caption{Comparison with baseline results of different generative methods on CT-RATE. All baseline results are directly cited from paper~\cite{ethemhamamci2025generatect}.}
  \resizebox{1\columnwidth}{!}{
  \begin{tabular}{p{0.01cm}lcccccc}
    \toprule
    &$\textbf{Method}$ & $\textbf{Out}$ & $\textbf{Time(s)}$   & $\textbf{FID}\downarrow$ & $\textbf{FVD}_{f16}\downarrow$ & \textbf{CLIP}$\uparrow$ \\
    \midrule
    \multirow{4}{*}{\rotatebox[origin=c]{90}{\textbf{Base}}}
    & w/ Imagen~\cite{ethemhamamci2025generatect}        & 2D       & 234    & 160.8   & 3557.7  & 24.8\\
    & w/ SD~\cite{ethemhamamci2025generatect}            & 2D       & 367    & 151.7   & 3513.5  & 23.5\\
    & w/ Phenaki~\cite{ethemhamamci2025generatect}       & 3D       & 23     & 104.3   & 1886.8  & 25.2\\
    & w/ GenerateCT~\cite{ethemhamamci2025generatect}    & 3D       & 184    & 55.8    & 1092.3  & 27.1\\
    \midrule
    \multirow{3}{*}{\rotatebox[origin=c]{90}{\textbf{Ours}}}
    &Fully-Body FMvT-L(StartBoost)  & 3D       & 99.37    & 67.91    & 683.02  & 16.67\\
    &GT-Head FMvT-L(StartBoost)  & 3D       & 94.97    & 37.10    & \textbf{618.86}  & 26.07\\
    &Next-Block FMvT-L(StartBoost)  & 3D       & 99.96    & \textbf{33.85}    & 978.96  & \textbf{41.22}\\
    \bottomrule
  \end{tabular}
  \label{tab:baseline-compare}
  }
\end{table}

\begin{table*}[ht]
  \centering
  \small
  \caption{Comprehensive comparison across inference styles, models, and resolutions.  We use three inference setups in our experiments. (1) The Full-Body method relies only on the text embedding. (2) The GT-Head generations are conditioned on both the text embedding, and relies in a real starting sequence (instead of the latent black sequence. Both of these methods (1 and 2) generate results iteratively, one sequence of 16 slices at a time. The Next-Block method generates each sequence directly from the previous real sequence, skipping the autoregressive process. The 512-resolution results are obtained by upscaling the 256-resolution images using bilinear interpolation. FMvT-L (StartBoost) indicates that the probability of sampling the first sequence (``start'') is increased to 30\% (``boost'') during training.}
  \resizebox{2.1\columnwidth}{!}{
  \begin{tabular}{llcccccccccc}
    \toprule
    \multicolumn{3}{c}{} &
    \multicolumn{4}{c}{\textbf{Resolution $256\times256$}} &
    \multicolumn{4}{c}{\textbf{Resolution $512\times512$}}\\
    \cmidrule(lr){4-7}\cmidrule(lr){8-12}
    \textbf{Inference} & \textbf{Model} & \textbf{Params (M)} &
    \textbf{FID}$\downarrow$ & $\textbf{FVD}_{f16}\downarrow$ & $\textbf{FVD}_{f128}\downarrow$ & \textbf{IS}$\uparrow$ & 
    \textbf{FID}$\downarrow$ & $\textbf{FVD}_{f16}\downarrow$ & $\textbf{FVD}_{f128}\downarrow$ & \textbf{IS}$\uparrow$ & \textbf{CLIP}$\uparrow$\\
    \cmidrule(lr){1-3}\cmidrule(lr){4-7}\cmidrule(lr){8-12}
    Full-Body & FMvT-S           &  36 & 249.73 & 2286.22 & 6646.07 & $2.29_{\pm0.18}$ & 245.18 & 2266.10 & 6634.59 & $2.32_{\pm0.18}$ & 11.71\\
              & FMvT-B           & 146 & 216.85 & 1861.98 & 2183.62 & $2.32_{\pm0.11}$ & 216.38 & 1810.07 & 2159.09 & $2.33_{\pm0.11}$ & 13.27\\
              & FMvT-L           & 512 & 131.30 & 1151.96 & 1220.31 & $2.21_{\pm0.13}$ & 130.51 & 1144.60 & 1209.47 & $2.25_{\pm0.14}$ & 9.93\\
              & FMvT-L (StartBoost) & 512 &  65.98 &  689.39 &  379.77 & $3.02_{\pm0.13}$ &  67.91 &  683.02 &  375.56 & $3.01_{\pm0.15}$ & 16.67\\
    \cmidrule(lr){1-3}\cmidrule(lr){4-7}\cmidrule(lr){8-12}
    GT-Head   & FMvT-S           &  36 & 151.15 &  623.23 & 3386.16 & $2.98_{\pm0.23}$ & 151.63 &  618.86 & 3401.43 & $2.95_{\pm0.19}$ & 23.11\\
              & FMvT-B           & 146 & 122.10 &  623.23 & 1183.50 & $2.98_{\pm0.23}$ & 112.72 &  618.86 & 1185.41 & $3.10_{\pm0.22}$ & 22.04\\
              & FMvT-L           & 512 &  67.46 &  623.23 &  725.81 & $2.35_{\pm0.20}$ &  65.04 &  618.86 &  715.64 & $2.40_{\pm0.19}$ & 21.64\\
              & FMvT-L (StartBoost)     & 512 &  36.91 &  623.23 &  389.66 & $2.71_{\pm0.23}$ &  37.10 &  618.86 &  374.77 & $2.71_{\pm0.14}$ & 26.07\\
    \cmidrule(lr){1-3}\cmidrule(lr){4-7}\cmidrule(lr){8-12}
    Next-Block & FMvT-S          &  36 & 116.94 & 914.29 & -- & $2.81_{\pm0.22}$ & 115.66 & 2265.92 & -- & $2.81_{\pm0.22}$ & 40.60\\
              & FMvT-B           & 146 &  69.12 & 456.46 & -- & $3.04_{\pm0.20}$ & 68.70  & 1809.19 & -- & $3.04_{\pm0.22}$ & 39.96\\
              & FMvT-L           & 512 &  37.50 & 200.06 & -- & $2.81_{\pm0.17}$ & 36.31  & 1143.65 & -- & $2.83_{\pm0.16}$ & 40.50\\
              & FMvT-L (StartBoost)     & 512 &  34.68 & 137.88 & -- & $2.88_{\pm0.24}$ & 33.85  & 678.96  & -- & $2.89_{\pm0.22}$ & 41.22\\
    \bottomrule
  \end{tabular}
  \label{tab:model-comparison}
  }
\end{table*}

\begin{table}[ht]
  \centering
  \small
  \caption{Comparison of different interpolation methods on the 256-resolution Full-Body FMvT-L experiment.}
  \resizebox{1\columnwidth}{!}{
  \begin{tabular}{llcccc}
    \toprule
    $\textbf{Resolution}$ & $\textbf{Method}$   & $\textbf{FID}\downarrow$ & $\textbf{FVD}_{f16}\downarrow$ & $\textbf{FVD}_{f128}\downarrow$ & \textbf{IS}$\uparrow$ \\
    \midrule
    256        & --       & 131.20    & 1115.63   & 1229.19  & $2.21_{\pm0.12}$\\
    512        & Bilinear & 130.51    & 1144.60   & 1209.47  & $2.25_{\pm0.14}$\\
    512        & Bicubic  & 132.66    & 1172.43   & 1228.74  & $2.17_{\pm0.12}$\\
    512        & Nearest  & 130.11    & 1148.71   & 1242.70  & $2.20_{\pm0.10}$\\
    \bottomrule
  \end{tabular}
  \label{tab:res-interp-compare}
  }
\end{table}

\subsection{Comparisons with State-of-the-art Methods}

\begin{figure}[t]
  \centering
  \includegraphics[width=1.0\linewidth]{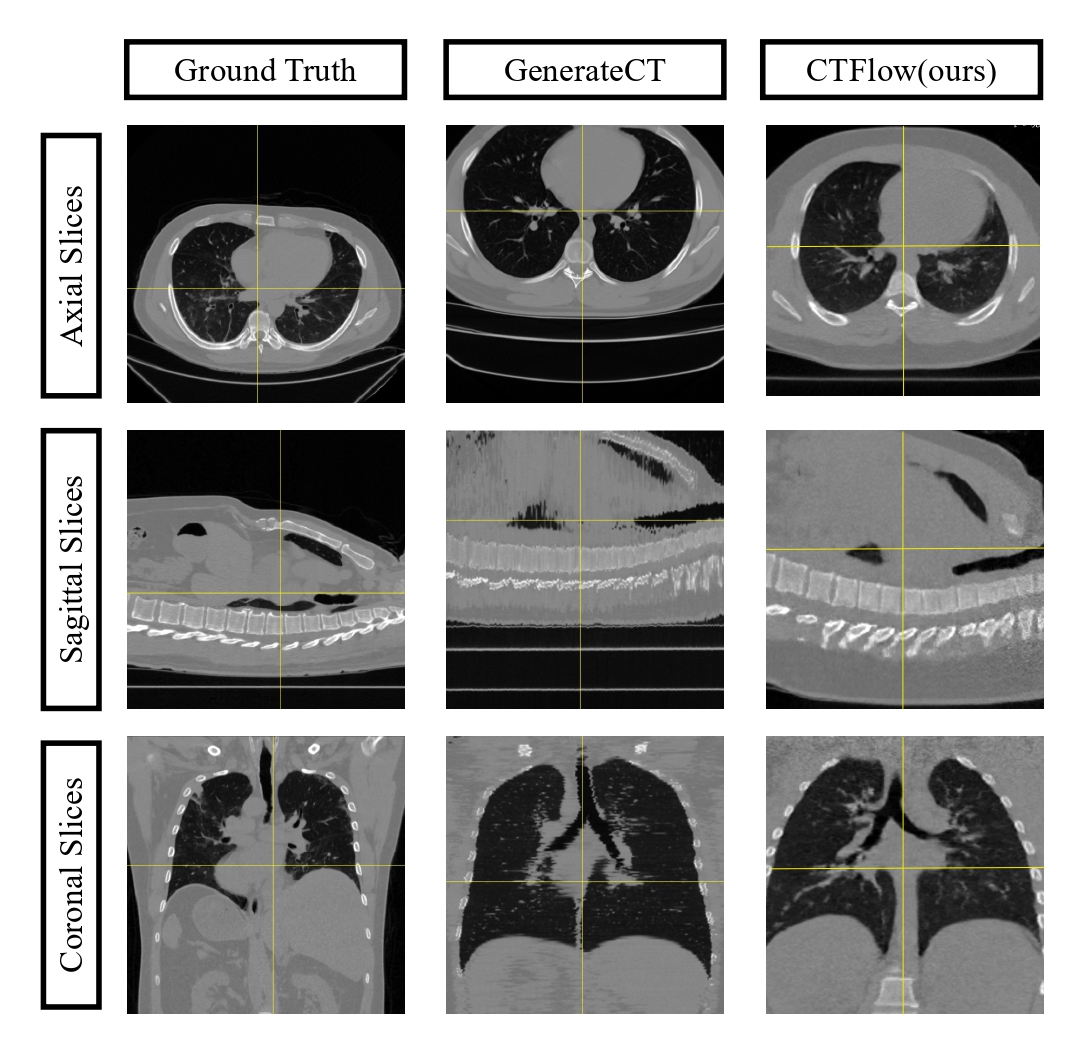}

   \caption{Axial, sagittal, and coronal slices of 3D volumes generated by various methods. From left to right, a real \gls{ct} volume, a synthetic \gls{ct} volume from~\cite{ethemhamamci2025generatect}, our synthetic volume using FMvT-L (StartBoost). Columns 1 and 2 are taken from~\cite{ethemhamamci2025generatect}.}
   \label{fig:3Dcompare}
\end{figure}

We benchmark our model against the state-of-the-art medical volume generation model, GenerateCT~\cite{ethemhamamci2025generatect}. We evaluate both models on the same CT-RATE validation set. As illustrated in Figure~\ref{fig:3Dcompare}, our model (third column) exhibits clear structural consistency. In sagittal and coronal views, our results demonstrate strong spatial coherence, while GenerateCT, due to its two-stage approach, produces blurrier results. 

Numerical results provide a more concrete demonstration of our model's performance. According to Hamamci et al.~\cite{ethemhamamci2024developing} and data in Table~\ref{tab:baseline-compare}, the best \gls{fid} achieved by GenerateCT is \textbf{55.8}, with a corresponding \gls{fvd} of \textbf{1092.3}, which serves as our baseline. 
From our experimental results in Table~\ref{tab:model-comparison} and Table~\ref{tab:baseline-compare}, we observe that the FMvT-L (StartBoost) model, under the fully-body inference setting, surpasses GenerateCT in terms of \gls{fvd}, indicating improved spatial coherence. 
Furthermore, our GT-head experiment shows that our model only shortcoming is the generation of the starting sequence. Indeed, the results of our GT-head experiment beat GenerateCT on both \gls{fid} and \gls{fvd} by a large margin, and match it on the CLIP score. 
Our next-block inference mode yields our best results, achieving an \gls{fid} as low as 34.68. 
The results at resolution $512 \times 512$ are comparable to those at the $256 \times 256$ original resolution. However, the interpolation introduces some blurring, resulting in slightly different metric values. 
All of these results indicate that our model is able to generate accurate and temporally coherent sequences (\emph{i.e.}, volumes) that closely follow the text conditioning.

Also, our CLIP score for the fully-body are very close to the best baseline results, and the CLIP score for the block-wise approach is almost twice as good as the baseline. This shows that our FMvT-L (StartBoost) model responds well to the information expressed in the text.

\subsection{Ablation Study}

\begin{figure}[t]
  \centering
  \includegraphics[width=1.0\linewidth]{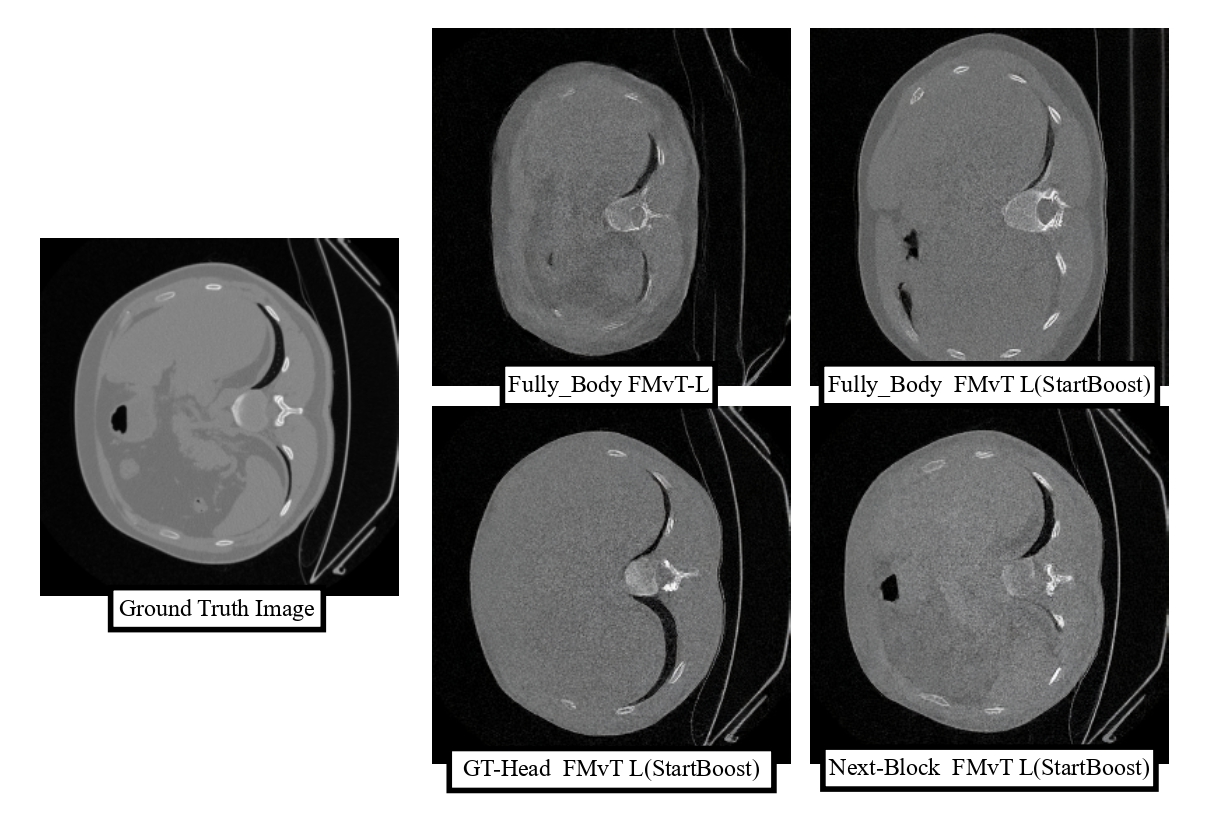}
   \caption{Visual Results Comparison Across Different Model Sizes}
   \label{fig:visual_result}
\end{figure}

As the original data has a slice resolution of $512 \times 512$, while our model is trained on a resolution of $256 \times 256$, we explore three interpolation methods that we apply as post-processing on our synthetic volumes. Results are presented in \Cref{tab:res-interp-compare}.
We observe that the choice of interpolation method has little impact on the final results, and that the results on the increased resolution remain within a few percentage points of the metrics on the original $256 \times 256$ resolution.
We ultimately employ bilinear interpolation, as it reaches the best score on all metrics.

Table~\ref{tab:model-comparison} also presents the performance of models of different sizes on the CT-RATE validation set. We can observe that as the model size increases, all evaluation metrics consistently improve: both \gls{fid} and \gls{fvd} decrease steadily, with \gls{fid}$_{f16}$ dropping from over 2286.22 to approximately 1151.96 in the fully-body setup. Furthermore, after increasing the sampling probability of the first clip, FMvT-L (StartBoost) outperforms all other models across all metrics. The \gls{is} score also increases from 2.29 (FMvT-S) to 3.02 (FMvT-L (StartBoost)), further validating the improvement in the diversity and quality of the generated results. These observations also indicate that when the model is trained with an emphasis on the very first sequence, its generation ability improves. Conversely, when the model is not accustomed to generating the first sequence, the bias tends to accumulate over the auto-regressive sampling, leading to compounding errors, hindering the fully-body generation results.

When the first sequence is excluded from evaluation (\emph{i.e.}, using GT-Head), the \gls{fvd} decreases and the \gls{is} increases notably compared to the fully-body setting, except for FMvT-L (StartBoost). This reinforces the observation that FMvT-L (StartBoost) substantially improves the quality of the first generated  sequence, thus reducing the bias introduced by generating the initial sequence. However, this bias is not completely eliminated, as the generated first sequence still lacks some details and realism. When the ground-truth first sequence is provided, the \gls{fid} drops considerably, further suggesting that generating the first sequence without any conditioning remains a challenging scenario.

Next-Block inference demonstrates the intrinsic generative capability of the model, as all 16-slice sequences are generated independently and conditioned only on the directly preceding real sequence. With this accurate conditioning, the model achieves the best metrics across the board, thanks to the higher quality of the conditioning it receives. This is also reflected in the CLIP scores, which demonstrate an even higher matching between the generated sequences and the text conditioning.


\section{Discussion and Limitations}
\label{sec:discussion}

Based on all ablation results, we evaluate our model in two core aspects: pure generation ability and iterative generation ability. Our results demonstrate that our model is capable of smoothly iterating over successive sequences, maintaining spatial coherence across the whole \gls{ct} volume. However, we observe that bias still tends to accumulate with each generated sequence, leading to a gradual loss of fine details as the sequence progresses. While conditioning on previous sequences improves temporal consistency, it also allows error propagation, which is an inherent problem to autoregressive approaches.

Unconditional generation of the first sequence remains the primary challenge. Without any conditioning information, the model often struggles to accurately capture the complexity and diversity of the initial segment. Future work will focus on developing more effective strategies for generating high-quality, unconditioned first sequences, thereby further improving overall volume quality and robustness.

\section{Conclusion}
\label{sec:conslusion}

We present a novel pipeline for \gls{ct} volume generation by modeling 3D medical volumes as sequences of 2D slices. By incorporating state-of-the-art video generation approaches into medical volume synthesis, our method achieves substantial improvements in spatial coherence. Furthermore, by operating in latent space with flow-matching, our model greatly reduces computational demands, making it more efficient for large-scale applications.

\noindent\textbf{Acknowledgments:} We acknowledge the Helmut Horten Foundation for their support, which made the CT-RATE dataset possible. We also sincerely thank Istanbul Medipol University Mega Hospital for their support and for providing the data. High-performance computing resources were provided in part by the Erlangen National High Performance Computing Center (NHR@FAU) at Friedrich-Alexander-Universität Erlangen-Nürnberg (FAU), under the NHR projects b143dc and b180dc. NHR is funded by federal and Bavarian state authorities, and NHR@FAU hardware is partially funded by the German Research Foundation (DFG) – 440719683. Additional support was  received by the ERC - project MIA-NORMAL 101083647,  DFG 513220538, 512819079, and by the state of Bavaria (HTA).

{
    \small
    \bibliographystyle{ieeenat_fullname}
    \bibliography{main}
}


\end{document}